%% file: dagstuhl.tex
\documentclass{article} 
\usepackage{amsmath}
\usepackage{amssymb}
\usepackage[dvips]{graphicx}
\usepackage{subfigure}
\usepackage{pstricks}
\usepackage{pst-node}
\usepackage{stmaryrd}
\usepackage{typearea}
\areaset{14.5cm}{22cm}

\include{macros}

\title{How to Evaluate Dimensionality Reduction?\\ -- Improving the Co-ranking Matrix}

\author{
Wouter Lueks$^{1,2}$, Bassam Mokbel$^1$, \\
Michael Biehl$^2$, Barbara Hammer$^1$ \\
\begin{small}
\begin{tabular}{cc}
																												  & \\
$^{1)}$ CITEC -- Center of Excellence											& $^{2)}$ Johann Bernoulli Institute\\
for Cognitive Interaction Technology										 	& for Mathematics and Computer Science \\
Bielefeld University 																			& University of Groningen \\
D-33501 Bielefeld 																				& P.O. Box 407, 9700 AK Groningen \\
Germany 																									& The Netherlands \\
\footnotesize{\texttt{\{wlueks|bmokbel|bhammer\}}} 				& \\
\footnotesize{\texttt{@techfak.uni-bielefeld.de}} 				& \footnotesize{\texttt{m.biehl@rug.nl}}
\end{tabular}
\end{small}
}

\date{
\vspace{0.5cm}
\begin{footnotesize}
{From Dagstuhl Seminar No. 11341: \\
\bf Learning in the context of very high dimensional data}\\
{21.08.11 - 26.08.11}\\[0.3cm]
Organizers: \\
Michael Biehl (Univ. of Groningen, NL), Barbara Hammer (Bielefeld Univ., DE), \\
Erzsébet Merényi (Rice Univ., US), Alessandro Sperduti (Univ. of Padova, IT), \\
Thomas Villmann (Univ. of Applied Sc. Mittweida, DE)
\end{footnotesize}
}

\begin{document}

\maketitle
\date{}

\begin{abstract}
The growing number of dimensionality reduction (DR) methods available for data visualization has recently inspired the development of quality assessment measures, in order to evaluate the resulting low-dimensional representation independently from a methods' inherent criteria.
Several (existing) quality measures can be (re)formulated based on the so-called co-ranking matrix, which subsumes all rank errors (i.e., differences between the ranking of distances from every point to all others, comparing the low-dimensional
representation to the original data).
The measures are often based on the partioning of the co-ranking matrix into 4 submatrices, divided at the K-th row and K-th column, calculating a weighted combination of the sums of each submatrix.
Hence, the evaluation process typically involves plotting a graph over several (or even
all possible) settings of the parameter K.
Considering simple artificial examples, we argue that this parameter controls
two notions at once, that need not necessarily be combined, and that the
rectangular shape of submatrices is disadvantageous for an intuitive
interpretation of the parameter.
We debate that quality measures, as general and flexible evaluation tools,
should have parameters with a direct and intuitive interpretation as to which
specific error types are tolerated or penalized.
Therefore, we propose to replace the parameter K with two distinct parameters to
control these notions separately, and introduce a differently shaped weighting
scheme on the co-ranking matrix.
The two new parameters can then directly be interpreted, respectively, as a threshold up to
which rank errors are tolerated, and a threshold up to which the rank-distances
are significant for the quality evaluation.
Moreover, we propose a color representation of local quality to visually support the
evaluation process for a given mapping, where every point in the mapping is colored according
to its local contribution to the overall quality value.
\end{abstract}
\newpage

\section{Introduction}
The amount of electronic data available today is becoming larger and larger in virtually all application domains;
at the same time, its complexity and dimensionality is increasing rapidly due to improved sensor technology and 
fine grained measurements as well as dedicated data formats.
In consequence, humans can no longer directly deal with such collections by inspecting the text files.
Rather, automated tools are required to support humans to extract the relevant information.
One core technology is given by data visualization: relying on one of the most
powerful senses of humans, it offers the possibility to visually inspect large amounts of data at once and to infer
relevant information based on the astonishing cognitive capabilities of humans in visual grouping
and similar.

Dimensionality reduction techniques constitute one important method in understanding high-di\-men\-sional data
because they directly produce a low-dimen\-sional visualization from high-dimensional vectorial data. 
Consequently, many dimensionality reduction techniques have been proposed in recent years. In the beginning, these methods were primarily 
linear, like principal component analysis (PCA), corresponding to low cost dimensionality reduction
techniques with a well founded mathematical background. However, linear techniques cannot preserve
relevant nonlinear structural elements of data.
Therefore, recently, more and more nonlinear methods like Isomap~\cite{Tenenbaum:2000}, locally linear embedding (LLE)~\cite{Roweis:2000}, and stochastic neighbor embedding (SNE)~\cite{Maaten:2008} have become popular, see the overview~\cite{DR_overview}, for example. 

With more and more dimensionality reduction techniques being readily available, the user faces the problem which of the methods
to choose for the current application. Usually, different techniques can lead to qualitatively very different results.
This is due to several reasons:
From a theoretical point of view, dimensionality reduction (DR) constitutes an ill-posed problem: Not all relations that exist in
the data in the high-dimensional space can be faithfully represented in the low-dimensional space, and it is not clear, which relations should be preserved. 
This has resulted in different objective functions for the various DR techniques corresponding to qualitatively very different
visualizations for a given data set.
In addition,
virtually all recent techniques have parameters to control the mapping.
These parameters usually control (in some way) which relations should be preserved best.
Hence, depending on the parameters of the method, even a single
DR technique can lead to qualitatively very diverse results.
In addition, many techniques do not lead to a unique solution, rather, multiple different outputs can be reached
due to random aspects of the algorithm corresponding to different local optima of the objective. Thus, it is even possible that qualitatively different results can be obtained by a single method with
a single set of model parameters. 

It is usually not clear whether the different results obtained by one or several
DR techniques correspond to different relevant
structural aspects in the data which are possibly partially contradictory
in low dimensions, or whether some of the methods and model parameters are less
suited to preserve the relevant structural aspects in the given data set.
Further, it can happen that suboptimal results are obtained simply because of numerical problems
of a given technique such as (bad) local optima.
At the same time, it is very hard for humans to judge the quality of a given mapping and the suitability
of a specific technique and choice of parameters by visual inspection:
the original data set is not accessible to the user due to its high dimensionality
such that a human cannot compare a given visualization to ground truth easily.
Therefore, there is a need to develop formal measures which judge the quality of a
given mapping of data. Such formal measures should evaluate, in an automated
and objective way, in how far structure of the original data corresponds to
the structure observed in the low-dimensional representation.

Formal measures to evaluate the quality of a data visualization have several benefits:
a formal measure directly  indicates whether an observer can trust a given mapping
or whether effects of the data visualization are caused by the choice of the method and parameters
rather than structures in the data set. Further, a formal quality measure can help humans
to choose a good method and a good set of parameters for a given data set. Often, 
it is not easy to predict the effects of model parameters on the resulting mapping. 
Good quality measures can serve as a means to interactively optimize the parameters. 
Last not least,  quality measures are generally important to automatically 
evaluate and compare DR techniques for research.

Several quality criteria to evaluate dimensionality reduction
have been proposed in recent years, see~\cite{Lee:2009} for an overview. 
As for dimensionality reduction itself, the problem to define formal evaluation
criteria suffers from the ill-posedness of the task:
it is not clear a priori which structural aspects of the data should
be preserved in a given task.
Most quality measures which have been proposed recently are based on
neighborhood relations of the data. They measure in some way in how far neighborhood relationships (e.g.\ ranks of data points) correspond to each other in the original space compared to the projection.
Two recent quality measures offer a general approach and constitute frameworks that include earlier measures as special cases~\cite{Lee:2009,Venna:2010}. 
Regarding this general framework, it becomes apparent that also the quality measure eventually depends on the needs of
the user since the user can specify, depending on the task, which aspects of the data are particularly relevant.
For example, when dealing with medical data, the user would prefer false positives over false negatives (whatever those may be in the case of dimensionality reduction). Also, data in which manifold structures are expected should be evaluated differently than clustered data.

Therefore, there is a need for intuitive and easily accessible quality measures
which allow the user to determine the precise form of the measure based on the current application.
That means, there is a need for quality measures with parameters which have
an intuitive meaning and which can easily be set by the user.
The co-ranking matrix~\cite{Lee:2009} already goes into this direction by pointing out the relevance
of the neighborhood rank which the user believes is important.

We will discuss that the global quality measure which has been derived based on this framework in the work~\cite{Lee:2009}
does not correspond to an intuitive interpretation by the user:
on the one hand, it depends on absolute values of the ranks rather than
the deviation of the ranks, i.e.\ the actual `errors' made by a DR method.
On the other hand, it relies on a single parameter only, the size of 
ranks taken into account, which controls both aspects: which errors are tolerated
and which neighborhood relations are considered interesting for the mapping.
We show in a simple example that this quality measure leads to unexpected values
which do not correspond to an intuitive understanding.

As an alternative, based on the co-ranking framework, we propose a different
family of quality criteria which are based on the values of the rank errors rather
than the absolute values of the ranks. This family is parameterized by two
parameters which control the size of errors which are tolerated on the one hand
and the size of the neighborhood of points which should be mapped faithfully by the dimensionality reduction
on the other hand.
This way, the user can intuitively control the resulting quality measure.
We also propose an intuitive way to link formal quality criteria to a given visualization
and to actually visualize the quality of a mapping such that the user can immediately see
which parts of the mapping are trustworthy and which are not.

\section{Dimensionality Reduction and Quality Measures\label{sect:objective_functions}}
Dimensionality reduction techniques are used for visualization by mapping a high-dimensional dataset $\hdimset = \set{\hdimpt{1},\ldots,\hdimpt{\nrpoints}}$ to a low-dimensional dataset $\ldimset = \set{\ldimpt{1},\ldots,\ldimpt{\nrpoints}}$. 
By design and via parameters, DR methods specify which properties should be maintained by the mapping. 
Some techniques are based on global mappings such as linear techniques, which
determine a matrix to reduce the dimensionality of the data set by a linear transformation,
or topographic mapping such as the self-organizing map \cite{Ultsch:1990}, which parameterize a mapping by 
a lattice of prototypes in the data space.
Many modern nonlinear techniques are non-parametric: they map a given set of data
points directly to their respective projections without specifying a functional form.
This way, the mapping has large flexibility and highly nonlinear effects can be obtained.

Non-parametric dimensionality reduction is often based on a cost function or objective, which
evaluates in how far characteristics of the original data $\hdimpt{i}$ are preserved by the
projections $\ldimpt{i}$. Appropriate projections are then determined minimizing
this objective with respect to the parameters $\ldimpt{i}$.
For example, t-SNE maintains the neighborhood probabilities in both spaces, while LLE tries to place points in such a way that locally linear neighborhoods are maintained. See e.g.\ the article \cite{IEEEpaper} for a general formalization of popular
non-parametric dimensionality reduction techniques in this way.

Let us look more closely at the LLE algorithm. First, one fixes a parameter $K$, that specifies the number of neighbors used to reconstruct a point. For each point $\hdimpt{i}$, find the $K$ closest neighbors and find weights $w_{ij}$ such that $\hdimpt{i}$ is best represented by $\sum_{j=1}^\nrpoints w_{ij}\hdimpt{j}$, where at most $K$ of the $w_{ij}$'s are allowed to be non-zero. Finally, low-dimensional mapped points $\ldimpt{i}$ are chosen such that the differences between $\ldimpt{i}$ and its neighbor-reconstruction $\sum_{j=1}^\nrpoints w_{ij}\ldimpt{j}$ are minimal.

A similar process description can be given for t-SNE~\cite{Maaten:2008}. First, using a fixed perplexity, the probabilities $\pi_{ij}$ that point $i$ in the high-dimensional space is neighbored to
point $j$ in the high-dimensional space are estimated. Similar probabilities $p_{ij}$ for the low-dimensional space are estimated using a Student-t distribution. The positions in the low-dimensional space are optimized by minimizing the Kullback-Leibler divergence between these distributions, i.e., minimizing their dissimilarity.

In both cases, a quality criterion, depending on a parameter, is used to specify an optimal mapping. 
Thus, for non-parametric DR methods, there is often a close relationship of
an objective function which in some way evaluates the quality of a mapping,
and a DR algorithm which actually finds projections such that the quality is optimized.
Of course, these criteria are partially chosen because they have a nice corresponding optimization method. 
For LLE, for example, a convex objective guarantees unique
solutions.
However, as demonstrated in the article~\cite{Venna:2010}, a quality criterion itself can also be used to derive a DR model
by means of a numeric optimization technique.
Thus, a quality criterion gives rise to a corresponding optimal mapping. 
Conversely, most existing non-parametric DR methods
induce a quality measure by which
a DR mapping can be evaluated.
Naturally, these evaluation criteria are highly biased towards the corresponding DR
technique. Further, they are usually not
parameterized in an intuitive way and they also incorporate aspects
which are important in order to obtain a numerically appealing optimization method rather
than a valid evaluation criterion only.

Here we are interested in a criterion which evaluates the quality of DR mappings in a uniform and intuitive
way, and which provides a parameterization that allows for an intuitive control by the user.
Thereby, it is irrelevant whether the resulting objective also leads to a simple optimization scheme.
First aproaches in this direction have been proposed based on the co-ranking framework
in the articles~\cite{Lee:2009,Lee:2010}.

\section{The Co-ranking Framework\label{sect:DR_and_QA}}
In the following, we introduce the co-ranking framework
as proposed by Lee and Verleysen~\cite{Lee:2009} and we recall several established quality criteria based thereon.
Let $\hdimdist{i}{j}$ be the distance from $\hdimpti$ to $\hdimptj$ in the high-dimensional space. 
Analogously, $\ldimdist{i}{j}$ is the distance from $\ldimpti$ to $\ldimptj$ in the low-dimensional space. 
From these distances we can compute the ranks of the neighbors for each point.
The rank of $\hdimptj$ with respect to $\hdimpti$ in the high-dimensional space is given by
\begin{equation*}
\hdimrank{i}{j} = \setsize{ \cset{k}{\hdimdist{i}{k} < \hdimdist{i}{j} \textrm{ or } (\hdimdist{i}{k} = \hdimdist{i}{j} \textrm{ and } 1 \leq k < j \leq \nrpoints )} },
\end{equation*} 
where $\setsize{A}$ is the cardinality of the set $A$,
resulting in the ranks
$\set{1,\ldots,\nrpoints - 1}$.
Analogously, the rank of $\ldimptj$ with respect to $\ldimpti$ in the low-dimensional space is given by
\begin{equation*}
\ldimrank{i}{j} = \setsize{ \cset{k}{\ldimdist{i}{k} < \ldimdist{i}{j} \textrm{ or } (\ldimdist{i}{k} = \ldimdist{i}{j} \textrm{ and } 1 \leq k < j \leq \nrpoints )} }.
\end{equation*}
Many existing quality criteria measure in how far ranks of points are preserved after the reduction to a low-dimensional space.
This way, local relationships are evaluated without referring
to irrelevant issues such as e.g.\ scaling of the data.

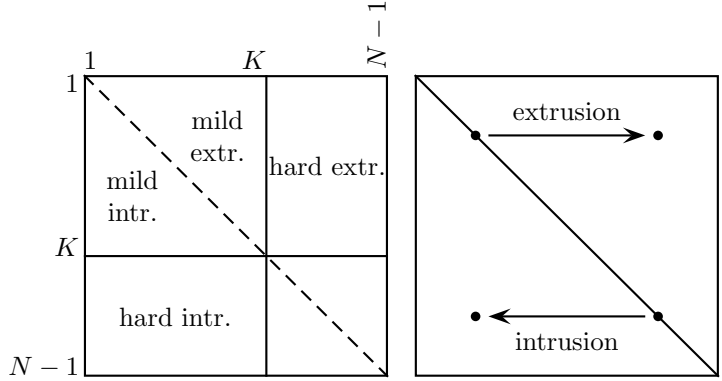
\begin{figure}
 \centering
 \subfigure{                    %
   \begin{pspicture}[showgrid=false](-1,0)(4,5) %
     \psset{xunit=0.8, yunit=0.8}
     \psframe(0,0)(5,5)
     \psline(0,2)(5,2)
     \psline(3,0)(3,5)
     \psline[linestyle=dashed](0,5)(5,0)

     \rput[c](4,3.5){hard extr.}
     \rput[c](1.5,1){hard intr.}

     \rput[c](0.8,3){\parbox{1cm}{\centering mild intr.}}
     \rput[c](2.2,4){\parbox{1cm}{\centering mild extr.}}

     \rput[lb](0,5.1){$1$} \rput[rt](-0.08,5){$1$}
     \rput[rb](3,5.1){$K$} \rput[rb](-0.08,2){$K$}
     \rput[lb]{90}(5,5.1){$\nrpoints - 1$} \rput[rb](-0.08,0){$\nrpoints - 1$}
   \end{pspicture}
 }
 \subfigure{                    %
 \begin{pspicture}[showgrid=false](0,0)(4,5)
     \psset{xunit=0.8, yunit=0.8}
     \psframe(0,0)(5,5)
     \psline(0,5)(5,0)


     \dotnode(1,4){A}
     \dotnode(4,4){B}
     \ncline[nodesep=3pt,arrowsize=6pt]{->}{A}{B} \naput{extrusion}

     \dotnode(4,1){C}
     \dotnode(1,1){D}           %
     \ncline[nodesep=3pt,arrowsize=6pt]{->}{C}{D} \naput{intrusion}
   \end{pspicture}
 }
 \caption{\label{fig:corankingmatrix} Large-scale structure of the co-ranking matrix. On the left, the matrix is split into blocks to show different types of intrusions and extrusions. In a perfect mapping, the co-ranking matrix will be a diagonal matrix. The image on the right shows how rank differences will alter the matrix. If a neighbor moves further away in the low-dimensional space, an extrusion, it will move mass to the right of the diagonal. Similarly, intrusions move mass to the left of the diagonal.}
\end{figure}

To generalize such measures,
the co-ranking matrix $Q$ \cite{Lee:2008} is defined by
\begin{equation*}
  Q_{kl} = \setsize{ \cset{(i,j)}{\hdimrank{i}{j} = k \textrm{ and }
      \ldimrank{i}{j} = l}}.
\end{equation*}
Errors of a DR mapping correspond to off-diagonal entries of this co-ranking matrix.
A point $j$ that gets a lower rank with respect to a point $i$ in the low-dimensional space than in the high-dimensional space, i.e.\ $\hdimrank{i}{j} > \ldimrank{i}{j}$, is called an \emph{intrusion}. Analogously, if $\hdimpt{j}$ has a higher rank in the low-dimensional space it is called an \emph{extrusion}. As shown in Figure~\ref{fig:corankingmatrix},
intrusions and extrusions correspond to off-diagonal entries in the upper or lower triangle, respectively.

Usually, a DR mapping is not used to map all relationships of data faithfully. Rather, the preservation of local
relationships is important. Hence, rank errors for large ranks are not as critical as rank errors
of close points. For this reason,
Lee and Verleyssen distinguish two types of intrusions/extrusions, those within a $K$-neighborhood, which are benevolent, and those moving across this boundary, which are malign with respect to quality.
A $K$-intrusion (resp.\ $K$-extrusion) is an intrusion for which $\ldimrank{i}{j} < K$ (resp.\ $\hdimrank{i}{j} < K$). Subsequently, mild $K$-intrusions are events for which $\ldimrank{i}{j} < \hdimrank{i}{j} \leq K$, while hard $K$-intrusions are defined by $\ldimrank{i}{j} \leq K < \hdimrank{i}{j}$. Mild $K$-extrusions and hard $K$-extrusions are defined accordingly.

Based on this setting, a simple quality measure can be defined:
it counts the number of points that remain inside the $K$-neighborhood while projecting, i.e., all points which keep their
rank, and all mild in- and extrusions:
\begin{equation}
\QNX{K} = \frac{1}{KN} \sum_{k=1}^K \sum_{l=1}^K Q_{kl}.
\label{quality}
\end{equation}
The normalization ensures that the quality of a perfect mapping equals one.\footnote{Instead of expressing the quality, one could define a measure of error analogously, as $1-\QNX{K}$.}

The quality criterion is very similar to the local continuity meta-criterion (LCMC) that was proposed by Chen and Buja \cite{Chen:2006}. In fact, it coincides up to a linear term that accounts for the quality of a random mapping:
\begin{equation}
\LCMC{K} =\QNX{K} - \frac{K}{N - 1}.
\end{equation}
Note that the range of this quality measure depends on $K$, i.e.\ the size of the neighborhood
which should be preserved by a DR mapping.
Often, a graph of the quality values over all possible $K$ (or a sufficient selection thereof) is plotted.
These measures have the disadvantage that they depend on $K$, and thus do not give a single decisive number that determines the quality of the mapping. 
Following the idea of locality-parameters in DR techniques (e.g., the number of neighbors in LLE and the perplexity in t-SNE) Lee and Verleysen make a similar local vs.\ global evaluation of the quality graph.

To estimate which values of $K$ should be considered local, the following splitting point was proposed in~\cite{Lee:2009}:
\begin{equation*}
  \Kmax = \arg\max_K \LCMC{K} = 
  \arg\max_K \left( \QNX{K} - \frac{K}{N-1} \right).
\end{equation*}
Subtracting the baseline ensures that there is a well-defined maximum that favors locality. 
Given the splitting point, a local and a global quality measure is obtained by averaging the respective parts of the quality graphs:
\begin{align}
  \Qlocal  &= \frac{1}{\Kmax} \sum_{K=1}^{\Kmax} \QNX{K}, \\
  \Qglobal &= \frac{1}{N - \Kmax} \sum_{K=\Kmax}^{N-1} \QNX{K}.
\end{align}
Both values range from 0 to 1. As local properties are more important, the authors advise to rank methods according to $\Qlocal$ and only to consider $\Qglobal$ in case of a tie.

\section{A family of quality criteria based on rank errors\label{sect:intuitive_control}}
The co-ranking matrix as introduced in~\cite{Lee:2009} offers a very elegant
framework to formalize quality criteria based on rank errors.
However, it has a severe drawback:
\emph{ 
The quality (\ref{quality}) depends on the number of rank errors in a
region of interest only, disregarding
the size of rank errors.}

Let us have a look at the evaluation measure (\ref{quality}).
A region of interest, i.e.\ a rank $K$ is fixed, following the idea that
ranks which are very large (larger than $K$) are not meaningful in the data space and
the projection space and thus, they can be disregarded.
Hence, only errors within the region of interest are counted.
The second role of $K$ is to define what is
regarded an error: an error occurs if and only if the region of interest in the
original space and the projection space does not coincide. Hence, the actual size of the rank error
is not important.
Rather, it is checked whether ranks $\le K$ keep this property while projecting and vice versa.
As an extreme case, points which change their rank from $1$ to $K$ are not
counted as errors, while points which change their rank only from $K$ to $K+1$ are.

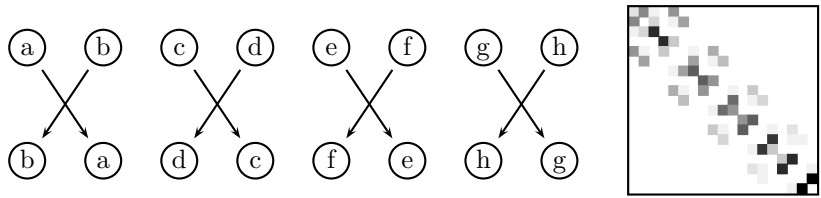
\begin{figure}[htb]
  \centering
\subfigure{
  \begin{pspicture}[showgrid=false](-0.3,-0.5)(7.5,1.8)
\Cnode(0,0){A}\rput[B](0,-0.1){b}
\Cnode(1,0){B}\rput[B](1,-0.1){a}
\Cnode(2,0){C}\rput[B](2,-0.1){d}
\Cnode(3,0){D}\rput[B](3,-0.1){c}
\Cnode(4,0){E}\rput[B](4,-0.1){f}
\Cnode(5,0){F}\rput[B](5,-0.1){e}
\Cnode(6,0){G}\rput[B](6,-0.1){h}
\Cnode(7,0){H}\rput[B](7,-0.1){g}

\psset{nodesep=3pt}
\Cnode(0,1.5){A2}\rput[B](0,1.4){a}
\Cnode(1,1.5){B2}\rput[B](1,1.4){b}
\ncline{->}{A2}{B}
\ncline{->}{B2}{A}

\Cnode(2,1.5){C2}\rput[B](2,1.4){c}
\Cnode(3,1.5){D2}\rput[B](3,1.4){d}
\ncline{->}{C2}{D}
\ncline{->}{D2}{C}

\Cnode(4,1.5){E2}\rput[B](4,1.4){e}
\Cnode(5,1.5){F2}\rput[B](5,1.4){f}
\ncline{->}{E2}{F}
\ncline{->}{F2}{E}

\Cnode(6,1.5){G2}\rput[B](6,1.4){g}
\Cnode(7,1.5){H2}\rput[B](7,1.4){h}
\ncline{->}{G2}{H}
\ncline{->}{H2}{G}

  \end{pspicture}
}
\subfigure{
  \begin{pspicture}[showgrid=false](0,0)(2.5,2.5)
  \psset{xunit=1.3mm, yunit=1.3mm, dotsize=1.3mm 1}
  
  \psset{dotstyle=square*, linecolor=black!10}

\psdot[linecolor=black!11](1,19)
\psdot[linecolor=black!47](2,19)
\psdot[linecolor=black!05](3,19)
\psdot[linecolor=black!42](5,19)
\psdot[linecolor=black!47](1,18)
\psdot[linecolor=black!16](3,18)
\psdot[linecolor=black!05](5,18)
\psdot[linecolor=black!37](6,18)
\psdot[linecolor=black!05](1,17)
\psdot[linecolor=black!16](2,17)
\psdot[linecolor=black!84](3,17)
\psdot[linecolor=black!84](4,16)
\psdot[linecolor=black!21](5,16)
\psdot[linecolor=black!42](1,15)
\psdot[linecolor=black!05](2,15)
\psdot[linecolor=black!21](4,15)
\psdot[linecolor=black!05](7,15)
\psdot[linecolor=black!32](9,15)
\psdot[linecolor=black!37](2,14)
\psdot[linecolor=black!37](7,14)
\psdot[linecolor=black!05](9,14)
\psdot[linecolor=black!26](10,14)
\psdot[linecolor=black!05](5,13)
\psdot[linecolor=black!37](6,13)
\psdot[linecolor=black!63](7,13)
\psdot[linecolor=black!63](8,12)
\psdot[linecolor=black!42](9,12)
\psdot[linecolor=black!32](5,11)
\psdot[linecolor=black!05](6,11)
\psdot[linecolor=black!42](8,11)
\psdot[linecolor=black!05](11,11)
\psdot[linecolor=black!21](13,11)
\psdot[linecolor=black!26](6,10)
\psdot[linecolor=black!58](11,10)
\psdot[linecolor=black!05](13,10)
\psdot[linecolor=black!16](14,10)
\psdot[linecolor=black!05](9,9)
\psdot[linecolor=black!58](10,9)
\psdot[linecolor=black!42](11,9)
\psdot[linecolor=black!42](12,8)
\psdot[linecolor=black!63](13,8)
\psdot[linecolor=black!21](9,7)
\psdot[linecolor=black!05](10,7)
\psdot[linecolor=black!63](12,7)
\psdot[linecolor=black!05](15,7)
\psdot[linecolor=black!11](17,7)
\psdot[linecolor=black!16](10,6)
\psdot[linecolor=black!79](15,6)
\psdot[linecolor=black!05](17,6)
\psdot[linecolor=black!05](18,6)
\psdot[linecolor=black!05](13,5)
\psdot[linecolor=black!79](14,5)
\psdot[linecolor=black!21](15,5)
\psdot[linecolor=black!21](16,4)
\psdot[linecolor=black!84](17,4)
\psdot[linecolor=black!11](13,3)
\psdot[linecolor=black!05](14,3)
\psdot[linecolor=black!84](16,3)
\psdot[linecolor=black!05](19,3)
\psdot[linecolor=black!05](14,2)
\psdot[linecolor=black!100](19,2)
\psdot[linecolor=black!05](17,1)
\psdot[linecolor=black!100](18,1)
\psframe[dimen=inner, linecolor=black](0.5,0.5)(19.5,19.5)
 \end{pspicture}
}
  \caption{\label{fig:mapping-swap} On the left is an example mapping from a one-dimensional set of equidistant points to a slight reordering. Since the points are only pairwise swapped, the changes in rank-distances are rather small. For the same setup with 20 points, this is confirmed by the co-ranking matrix that is shown on the right, with entries depicted on a gray scale. White indicates a zero value, while black corresponds to the maximum value in the matrix.}
\end{figure}

\begin{figure}[htb]
	\centering
	\subfigure[\label{fig:switchedrow_Qnx}Quality $\QNX{K}$] {
		\includegraphics[width = 0.55\linewidth]{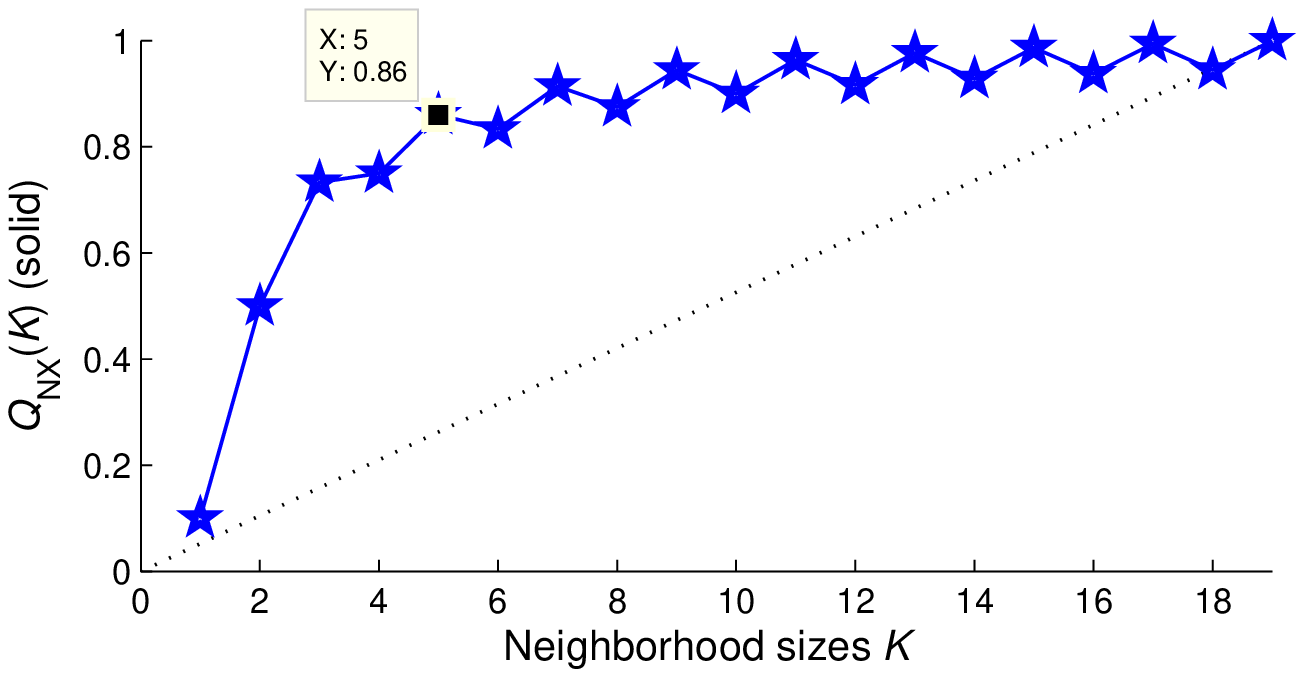}
	}
	\subfigure[\label{fig:switchedrow_Qnew}Quality $\Qnew{\kappa_s}{\kappa_t}$] {
		\includegraphics[width = 0.40\linewidth]{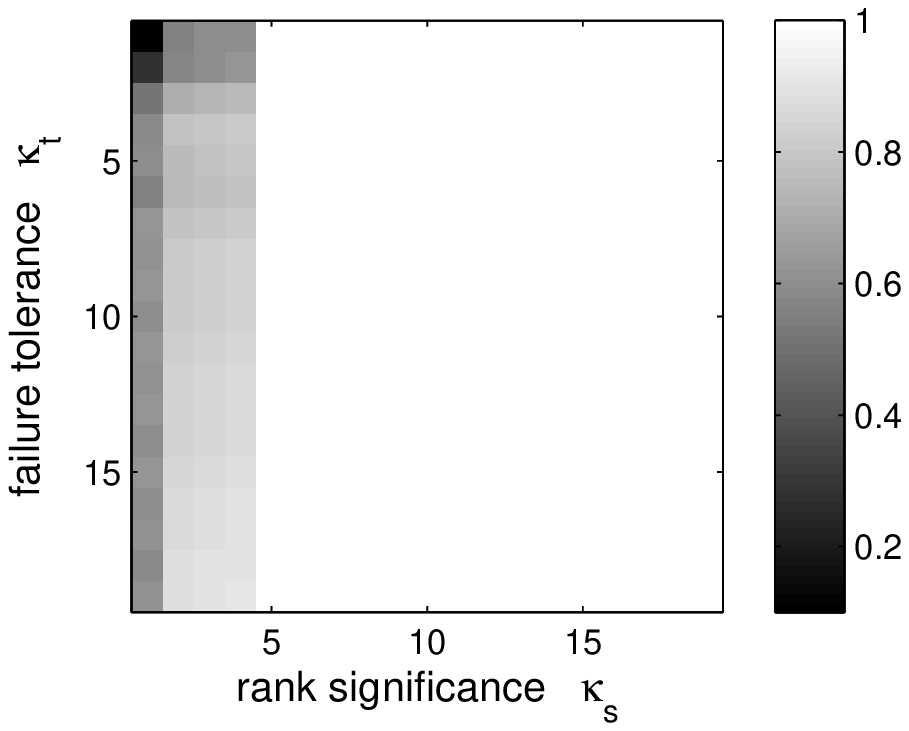}
	}
	\caption{The figure shows quality evaluations for the mapping presented in Figure~\ref{fig:mapping-swap}, on the left with the established measure $\QNX{K}$, on the right with the proposed new measure $\Qnew{\kappa_s}{\kappa_t}$. Both are evaluated for all possible parameter settings $K$ and $(\kappa_s,\kappa_t)$ respectively. The particular position of $K=5$ is highlighted on the graph in the left figure. $\QNX{K}$ with $K\ge5$ does not yield values of 1 entirely, which seems rather unintuitive for the given mapping. As expected, the matrix for $\Qnew{\kappa_s}{\kappa_t}$ does have ones, for all $\kappa_s\ge5$.}
\end{figure}

This definition of $\QNX{K}$ can lead to curious situations, which demonstrate the unintuitive character of 
the parameter $K$. Consider the pairwise swapping of points that is shown in Figure~\ref{fig:mapping-swap}. The number of points can be chosen arbitrarily. Examining the structure quickly shows that the maximum rank error between these permutations is at most 4 (for example, when the base point moves left, and the other point moves right)\footnote{Note that the points are equidistant, and in case of a tie in distances, the point with the lower alphabetical letter gets assigned the lower rank.}. Intuitively, if we consider rank error sizes up to 4 as acceptable, this mapping is perfect. This is, however, not the case when looking at $\QNX{5}$: there are still errors. In fact, for every value of $K$ there will be some point that moves from, for example, rank $K$ to a slightly higher rank, and therefore be counted as an error. This is also confirmed by the graph in Figure~\ref{fig:switchedrow_Qnx} which displays the quality for a row of 20 points.
It is hardly possible to detect an intuitive correlation of the size of $\QNX{K}$ and the parameter $K$
based on the observation of the mapping, albeit its errors can be fully characterized
by local pairwise swappings.

A look at the co-ranking matrix in Fig.~\ref{fig:mapping-swap} indicates the underlying structure in this case.
Since the rank error is always smaller than $5$, only $4$ off-diagonals of the co-ranking matrix
are non-vanishing. Note that the $i$th off-diagonal corresponds to rank errors of size
$i$. The quality measure~(\ref{quality}), however, only sums over rectangular parts of the 
co-ranking matrix. 
This observation also suggests how the quality~(\ref{quality}) can be altered to 
lead to a more intuitive parameterization:
rather than rectangular parts only, it should focus on a limited number of
off-diagonals corresponding to the size of the rank deviation which is considered acceptable.

As already mentioned, $K$ has also a different role by singling out the region of interest
based on the rationale that large ranks are not relevant anyway.
For the original quality~(\ref{quality}), this
choice is controlled by the same parameter $K$: points up to rank $K$ only are considered 
relevant, at the same time leading to the consequence that errors within this range $K$
are fully accepted.
It seems more intuitive to separating this control parameter:
in addition to the size of the errors which are tolerated, a region of interest
should be controlled independently.

Now, we will formalize this consideration by first reformulating the quality~(\ref{quality})
such that the two different roles of $K$ become apparent, and
then generalizing this formalization such that an explicit control
of the region of interest and the tolerated rank error size becomes possible.

\begin{figure}
 \centering
 \subfigure{                    %
   \begin{pspicture}[showgrid=false](-1,0)(4,5) %
     \psset{xunit=0.8, yunit=0.8}
     \psframe(0,0)(5,5)
     \psline(0,2)(5,2)
     \psline(3,0)(3,5)
     \psline[linestyle=dashed](0,5)(5,0)

     \rput[c](4,3.5){hard extr.}
     \rput[c](1.5,1){hard intr.}

     \rput[c](0.8,3){\parbox{1cm}{\centering mild intr.}}
     \rput[c](2.2,4){\parbox{1cm}{\centering mild extr.}}

     \rput[lb](0,5.1){$1$} \rput[rt](-0.08,5){$1$}
     \rput[rb](3,5.1){$K$} \rput[rb](-0.08,2){$K$}
     \rput[lb]{90}(5,5.1){$\nrpoints - 1$} \rput[rb](-0.08,0){$\nrpoints - 1$}
   \end{pspicture}
 }
\subfigure{
 \begin{pspicture}[showgrid=false](-1,0)(4,5)
     \psset{xunit=0.8, yunit=0.8}

\pspolygon*[linecolor=lightgray,dimen=inner](0,4)(3,1)(3,2)(4,2)(1,5)(0,5)
     \psframe(0,0)(5,5)
     \psline[linestyle=dashed]{cc-cc}(0.1,4.9)(4.9,0.1)

     \psframe(3,0)(5,2)
     \psline{cc-c}(0,4)(3,1)
     \psline{cc-c}(1,5)(4,2)

     \rput[rB](-0.08,3.9){$\kappa_t$}
     \rput[b](1,5.1){$\kappa_t$}

     \rput[t](2.5,.5){$\kappa_s$}
     \rput[Bl](5.1,1.9){$\kappa_s$}
 \end{pspicture}
 }
 \caption{New weighting scheme for the co-ranking matrix (right) as
 compared to the one proposed by Lee and Verleysen \cite{Lee:2009} (left):
 as before, the region of interest is limited to points which
 have a rank smaller than $\kappa_s$ either in the original space or the projection space, i.e.\
 the lower right corner is not considered. Within this region of interest,
 all pairs of points with a rank error smaller than a given tolerance
 $\kappa_t$ are counted as correctly projected.}\label{fig:diagonal}
\end{figure}
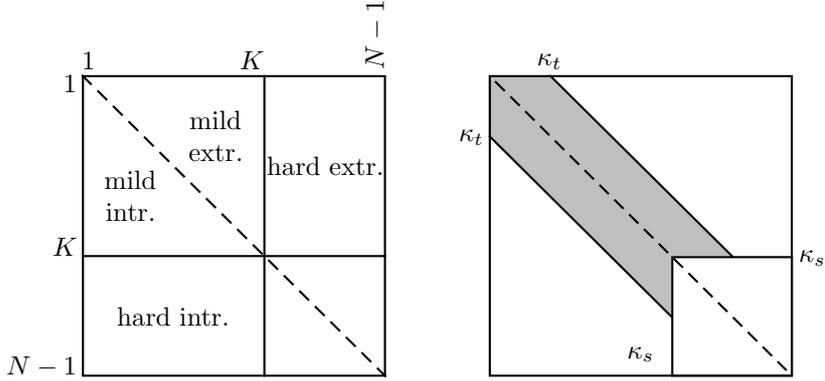 

\subsection{Analysis of $\QNX{K}$}
As discussed in~\cite{Lee:2009}, formal evaluation measures judge in how far 
ranks of points are preserved. Hence, the evaluation measures are based on rank errors
\begin{equation*}
  E_{ij} = \lvert \hdimrank{i}{j} - \ldimrank{i}{j} \rvert
\end{equation*}
which measure the deviation of the rank of two points in the original space and the
projection space, respectively.
The parameter $K$ has two distinct functions: 
it limits the region of interest, i.e.\ the relevant ranks which are regarded when evaluating the quality;
and it specifies what is regarded as an error, i.e.\ which rank
deviations are tolerated.

Formally, the first role of $K$
can be captured by a \emph{rank significance} function $\weightsignificance: R \times R \times \mathbb{N} \to [0,1]$ 
that determines, for any pair of points $i$ and $j$ the extent $\weightsignificance(\hdimrank{i}{j},\ldimrank{i}{j},K)$ to which their rank error should be taken into account. For the quality measure $\QNX{K}$, we have: 
\begin{equation*}
  \weightsignificance(\hdimrank{i}{j},\ldimrank{i}{j},K) =
  \left\{
    \begin{array}{ll} 
      0 & \hdimrank{i}{j} > K \land \ldimrank{i}{j} > K \\
      1 & \textrm{otherwise.}
    \end{array}
  \right.
\end{equation*}

To describe the second role of $K$,
we use a function $\weighttolerance: R \times R \times \mathbb{N} \to [0,1]$ that determines the weight of the rank error $E_{ij}$ for points $i$ and $j$ based on their ranks $\hdimrank{i}{j}$ and $\ldimrank{i}{j}$. 
According to the definition of $\QNX{K}$, we have the following \emph{error tolerance} function:
\begin{equation*}
  \weighttolerance(\hdimrank{i}{j},\ldimrank{i}{j},K) =
  \left\{
    \begin{array}{ll} 
      0 & (\hdimrank{i}{j} \leq K \land \ldimrank{i}{j} > K) \lor
          (\hdimrank{i}{j} > K \land \ldimrank{i}{j} \leq K)\\
      1 & \textrm{otherwise.}
    \end{array}
  \right.  
\end{equation*}
where we count the overlap of the $K$ neighborhoods in the original space and the projection space, respectively.
The quality is proportional to the number of points which are in the region of interest and which error is acceptable according to
the error tolerance:
\begin{equation}
\label{QNX-splitted}
  \QNX{K} = \frac{1}{K \nrpoints} \sum_{i=1}^{\nrpoints} \sum_{j=1}^{\nrpoints}
    \weightsignificance(\hdimrank{i}{j}, \ldimrank{i}{j}, K) \cdot
    \weighttolerance(\hdimrank{i}{j}, \ldimrank{i}{j}, K).
\end{equation}

As discussed before, a problem of the error tolerance function $\weighttolerance$ is that this function depends on the actual ranks and not 
on the rank error. 
Examining Figure~\ref{fig:corankingmatrix} confirms this. A point with high-dimensional rank 1 and low-dimensional rank $K$ is acceptable, although it has an absolute rank error of $K - 1$. On the other hand, a point that has high-dimensional rank $K$ and low-dimensional rank $K + 1$ is not acceptable, although its rank error is only $1$.

\subsection{A quality measure based on rank errors\label{sect:rank_independence}}

Because of this fact, we propose the following alternative error tolerance function
\begin{equation*}
  \weighttolerance(\hdimrank{i}{j},\ldimrank{i}{j},\kappa_t) =
  \left\{
    \begin{array}{ll} 
      0 & \lvert \hdimrank{i}{j} - \ldimrank{i}{j}\lvert \ > \kappa_t \\
      1 & \textrm{otherwise,}
    \end{array}
  \right.
\end{equation*}
that depends on the rank error rather than the value
of the ranks. The cut-off value $\kappa_t$ determines which error sizes are accepted. 
We use the same rank significance function $\weightsignificance$ as in~(\ref{QNX-splitted}), but substitute
the parameter $K$ by the the cut-off parameter $\kappa_s$.
Following equation~\eqref{QNX-splitted}, we then get a new quality measure:
\begin{equation}
\Qnew{\kappa_s}{\kappa_t} = \frac{1}{\kappa_s \nrpoints} 
  \sum_{i=1}^N \sum_{j=1}^N 
    \weightsignificance(\hdimrank{i}{j}, \ldimrank{i}{j}, \kappa_s) \cdot
    \weighttolerance(\hdimrank{i}{j}, \ldimrank{i}{j}, \kappa_t).
\end{equation}
Because of the normalization, values are in the interval $[0,1]$
with $1$ corresponding to a perfect mapping.
Figure~\ref{fig:diagonal} shows the region of the co-ranking matrix which is taken into account in this quality measure.
One might also consider more complex or smooth functions for $\weightsignificance$ and $\weighttolerance$ than simple cut-offs with $\kappa_s$ and $\kappa_t$ respectively and corresponding normalization factors,
to account for a smooth weighting of the neighborhood size and the size of the errors..

The new quality measure $\Qnew{\kappa_s}{\kappa_t}$ now depends on two parameters instead of only one $K$
which allow for an intuitive access:
$\kappa_t$ determines which size of the rank errors are tolerated,
while $\kappa_s$ singles out which ranks fall into the region of interest.

We can no longer plot only one graph to picture the quality of a mapping at different stages of $K$.
To get a rich impression of a mapping's qualitative characteristics, one can instead display a 3D surface
or colored matrix, where the position $(\kappa_s,\kappa_t)$ is assigned the value $\Qnew{\kappa_s}{\kappa_t}$, see Figure~\ref{fig:switchedrow_Qnew} for an example.
The matrix in Figure~\ref{fig:switchedrow_Qnew} shows all values of $\Qnew{\kappa_s}{\kappa_t}$ for the example in Figure~\ref{fig:mapping-swap}.
It clearly shows that the maximum quality is reached for all $\kappa_s > 4$.

Similar to \cite{Lee:2009}, it is possible to subtract the baseline given by a random mapping.
Based on this `centered' graph, a baseline can be determined up to which border local values in the matrix are interesting. Then, the given matrix can be connected to a single scalar by
averaging over these numbers.

%
%
%
%

\section{Local quality Assessment\label{sect:local_quality}}

The quality criteria introduced in the previous section average the
contributions of all points.
It can be useful to visually represent the quality (or analogously error) of a single point, in order to gain insight into local qualitative changes, especially when the deviation 
of the quality of the mapping in the single parts is very high.
This principle has been used, for example, to visualize the topographic distortion of self-organizing maps, where one can display the distance between neurons in the data space as a color in the topographic map, see~\cite{Ultsch:1990}. Similarly,
in the approach \cite{aupetit}, the local topographic reliability of dimensionality reduction
is displayed. 

The quality measure as introduced above
naturally gives rise to a local quality which, given a single point, displays the
trustworthiness of the map in this area.

\begin{figure}[tb]
  \centering
  \subfigure[Simple example]{
  \begin{pspicture}[showgrid=false](0.7,0.3)(5.0,1.7)
\Cnode(1,1.5){A}\rput[B](1,1.4){a}
\Cnode(2,1.5){B}\rput[B](2,1.4){b}
\Cnode(4,1.5){C}\rput[B](4,1.4){c}

\Cnode(1,0.5){C2}\rput[B](1,0.4){c}
\Cnode(2,0.5){B2}\rput[B](2,0.4){b}
\Cnode(4,0.5){A2}\rput[B](4,0.4){a}
  \end{pspicture}
  }
  \subfigure[High- and low-dimensional ranks]{
    \begin{tabular}[b]{c|ccc} 
$\hdimrank{\shortdownarrow}{\shortrightarrow}$ & a & b & c \\
\hline
    a &   & 1 & 2 \\
    b & 1 &   & 2 \\
    c & 2 & 1 &   \\
    \end{tabular} $\quad$

    \begin{tabular}[b]{c|ccc}
$\ldimrank{\shortdownarrow}{\shortrightarrow}$ & a & b & c \\
\hline
    a &   & 1 & 2 \\
    b & \colorbox{lightgray}{2} &   & \colorbox{lightgray}{1} \\
    c & 2 & 1 &   \\
    \end{tabular}
  }
  \caption{\label{fig:rankerrors} The items a and b are interchanged in the one-dimensional dataset shown on the left. This results in the rank-matrices as shown on the right, errors are highlighted.}
\end{figure}
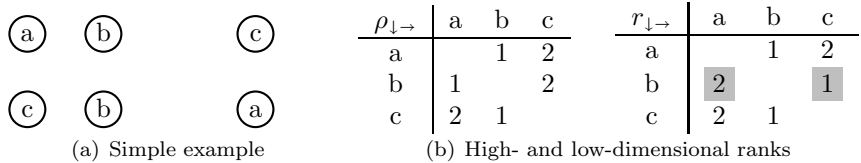

First, we examine what to count as visualization errors. To ease this, we assume that every rank change is counted as an error for now, i.e. $\kappa_t = 1$ and $\kappa_s = \nrpoints-1$. As a first attempt, one might choose the following point-wise error measure for point $i$:
\begin{equation*}
Q^{\prime}_i = \frac{1}{\kappa_s N} \sum_{j=1}^\nrpoints
		\weightsignificance(\hdimrank{i}{j}, \ldimrank{i}{j}, \kappa_s) \cdot
    \weighttolerance(\hdimrank{i}{j}, \ldimrank{i}{j}, \kappa_t).
\end{equation*}
However, this is not optimal. 
Consider the simple example shown in Figure~\ref{fig:rankerrors}. Interchanging the elements a and c induces only rank changes for point b. Given the formula $Q^\prime_i$, only point b would be indicated as erroneous, which does not really reflect our intuitive notion of error. Similarly, we can assume fixed positions for a and c, and move b towards c. Then the ranks for a and c change, while b's remain unchanged, and hence b gets no error. Therefore, we propose to use the symmetric quality:
\begin{alignat}{1}
Q^{\prime}_i = \frac{1}{2 \kappa_s N} \sum_{j=1}^\nrpoints 
			\Big(
			\weightsignificance(\hdimrank{i}{j}, \ldimrank{i}{j}, \kappa_s) \cdot
   		\weighttolerance(\hdimrank{i}{j}, \ldimrank{i}{j}, \kappa_t) \Big. \ + \label{eq:Qi}\\[-3pt]
   		\Big.
			\weightsignificance(\hdimrank{j}{i}, \ldimrank{j}{i}, \kappa_s) \cdot
    		\weighttolerance(\hdimrank{j}{i}, \ldimrank{j}{i}, \kappa_t) 
\Big). \notag
\end{alignat}
This choice assigns a reduced quality to all the three points in the example of Figure~\ref{fig:rankerrors}.
Given this measurement, it is easy to color every point in a low-dimensional data visualization according to
the local quality measured at the point.

\begin{figure}[htb]
	\centering
	\subfigure[\label{fig:swissroll_original}Original in $\mathbb{R}^3$] {
		\includegraphics[width = 0.25\linewidth]{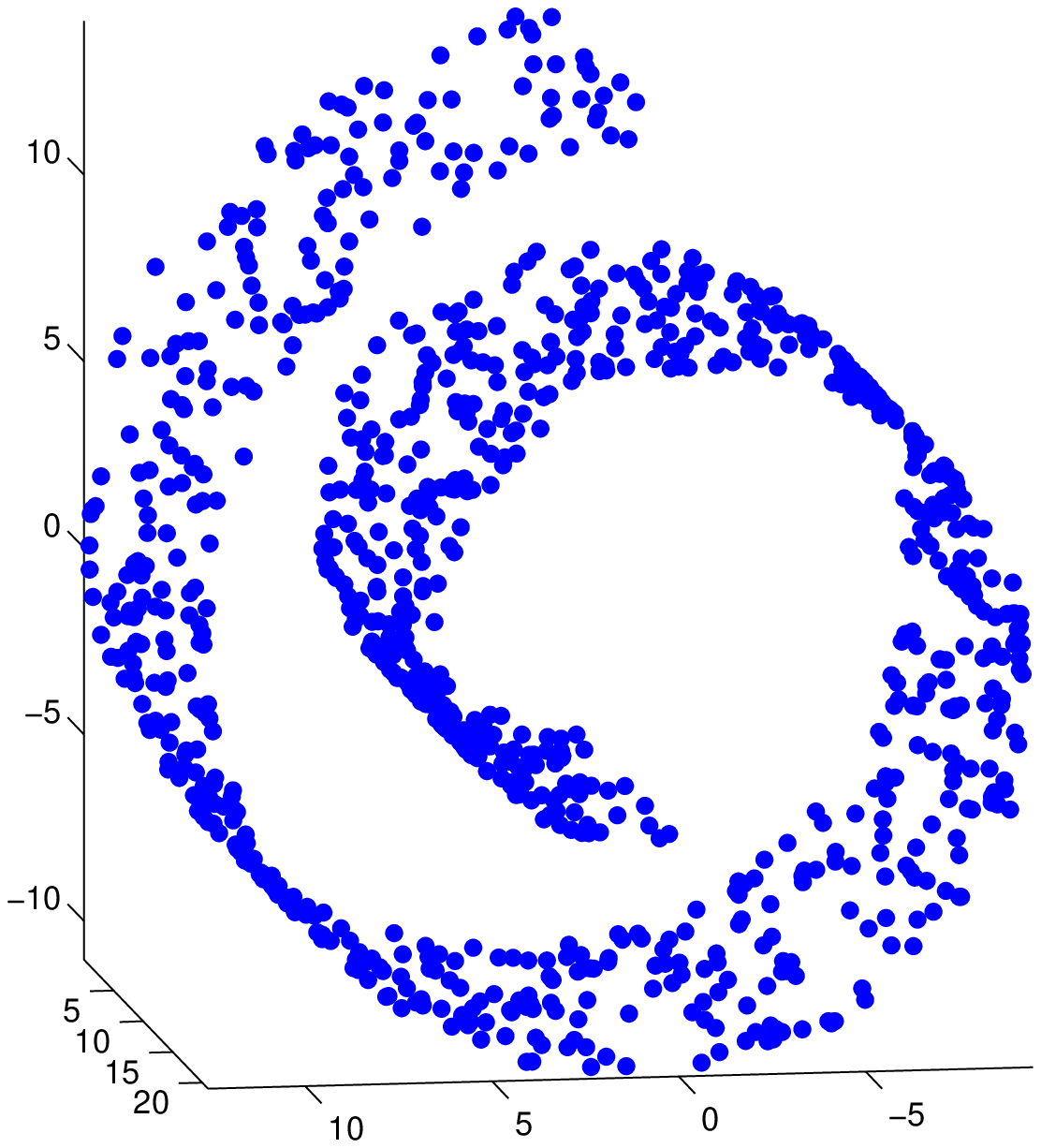}
	}
	\subfigure[\label{fig:swissroll_colored}t-SNE mapping, colored by pointwise quality contributions] {
		\includegraphics[width = 0.7\linewidth]{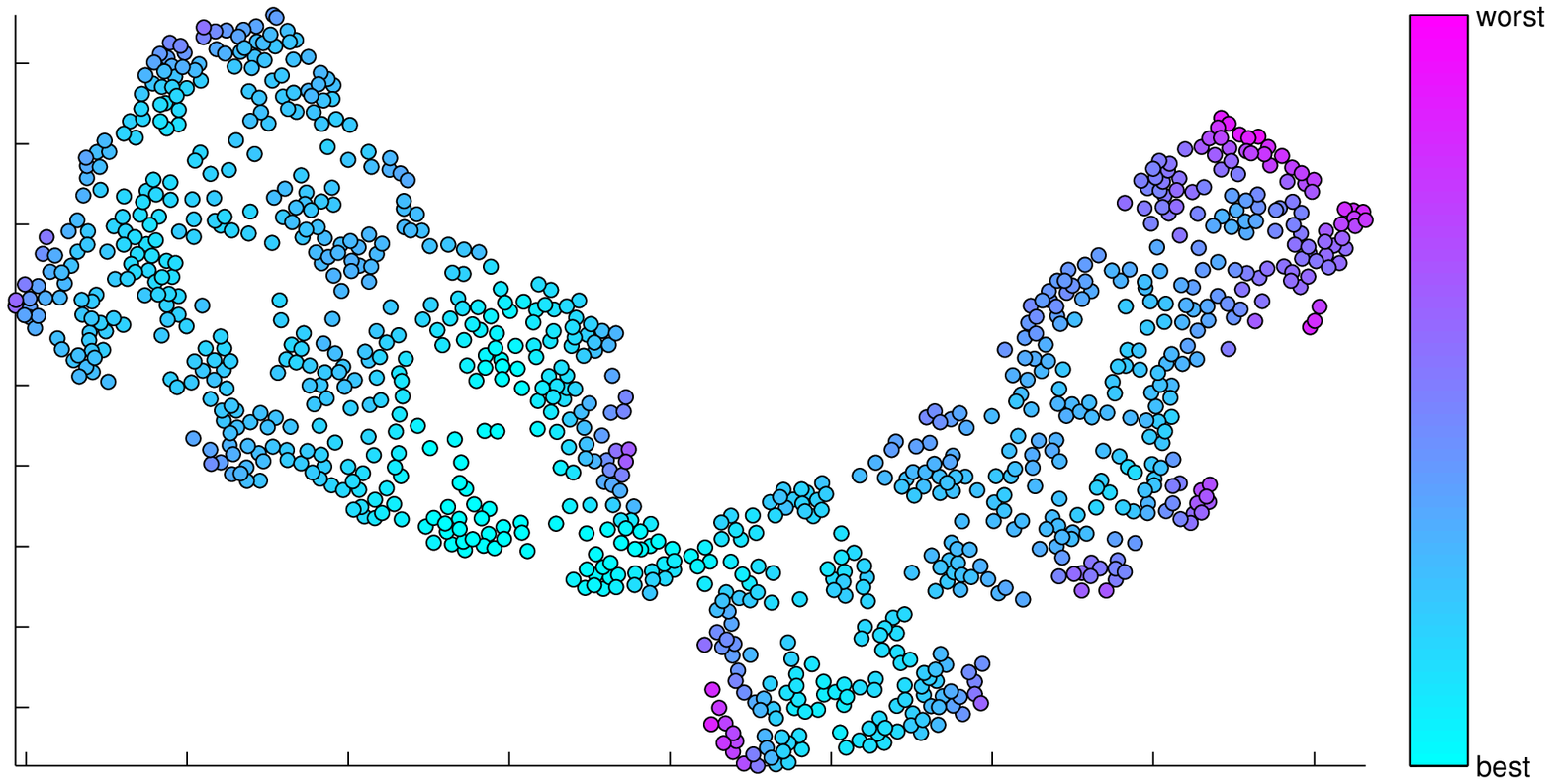}
	}
	\caption{\label{fig:swissroll_coloring}The left figure shows a view of the swiss roll benchmark data set in its original three dimensions, the right-side picture shows a two-dimensional embedding of t-SNE, using a perplexity parameter of 50. Every data point is colored by its amount of contribution to the overall quality, see Equation~(\ref{eq:Qi}), with $\kappa_s=96$ and $\kappa_t=70$. Altogether, the visualization by t-SNE seems to be appropriate for the given data, since the global manifold structure is largely maintained. This is confirmed by the quality-coloring. However, the coloring reveals local tearing of the manifold, as well as errors on the right border, caused by 'unrolling' the inner end of the spiral.
	The latter occurs when referring to the standard Euclidean distance in the original space.
	Taking geodesic distances, this effect vanishes and only the part where the manifold is teared would be highlighted.
	The overall quality evaluation is depicted in Figure~\ref{fig:swissroll_qualities}.}
\end{figure}

We show an example where this simple
coloring scheme clearly indicates 
which regions of a mapping can be
considered trustworthy.
In Figure~\ref{fig:swissroll_original}, we show the popular 'swiss roll' benchmark data set, which consists of points in $\mathbb{R}^3$, uniformly sampled on a spiral-shaped two-dimensional manifold.
The data is mapped by t-SNE using a high perplexity parameter which produces an 'unfolded' view of the manifold, with some local tearing and distortion, see Figure~\ref{fig:swissroll_colored}.
The coloring clearly reveals the tears within the manifold as well as the larger rank errors that occur at the rightmost points caused by 'unrolling' and putting the inner end of the belt far away from its original neighbors on the next spiral loop level. 
In a real world scenario, where the original data is high-dimensional and its detailed structure is unknown to the user, the coloring of the mapped points may help to understand local characteristics of the mapping.

\begin{figure}[htb]
	\centering
	\subfigure[\label{fig:swissroll_Qnx}Quality $\QNX{K}$] {
		\includegraphics[width = 0.8\linewidth]{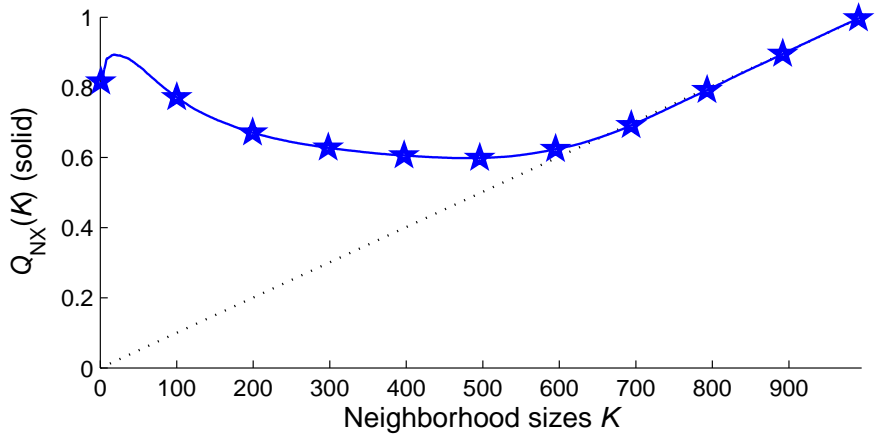}
	}\\
	\subfigure[\label{fig:swissroll_Qnew}Quality $\Qnew{\kappa_s}{\kappa_t}$] {
		\includegraphics[width = 0.8\linewidth]{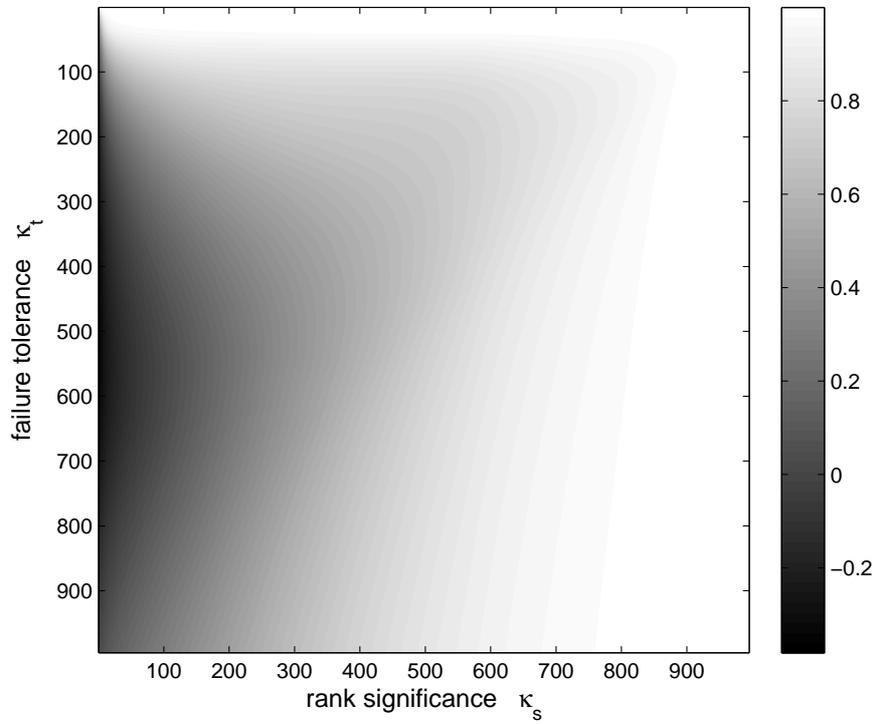}
	}
	\caption{\label{fig:swissroll_qualities}The figure shows quality evaluations for the mapping presented in Figure~\ref{fig:swissroll_coloring}; on top with the established measure $\QNX{K}$, on the bottom with the proposed new measure $\Qnew{\kappa_s}{\kappa_t}$. Both are evaluated for all possible parameter settings $K$ and $(\kappa_s,\kappa_t)$ respectively. We recommend to read the bottom matrix by first focusing on some error tolerance value $\kappa_t$, and then read one row of the matrix with all rank significance values $\kappa_s$\,.}
\end{figure}

\begin{footnotesize}
\bibliographystyle{unsrt}
\bibliography{bibliography}
\end{footnotesize}

\end{document}

%% file: macros.tex
\newcommand{\setsize}[1]{\lvert #1 \rvert}
\newcommand{\set}[1]{\ensuremath{ \left\{ #1 \right\}}}
\newcommand{\cset}[2]{\ensuremath{ \left\{ #1 \mid #2 \right\} }}

\newcommand{\LCMC}[1]{\textrm{LCMC}(#1)}
\newcommand{\QNX}[1]{Q_{\textrm{NX}}(#1)}

\newcommand{\Qnew}[2]{Q_{\textrm{NX}}^\prime(#1,#2)}

\newcommand{\Kmax}{K_{\textrm{max}}}

\newcommand{\Qlocal}{Q_{\textrm{local}}}
\newcommand{\Qglobal}{Q_{\textrm{global}}}

\newcommand{\weightsignificance}{w_s}
\newcommand{\weighttolerance}{w_t}

\newcommand{\nrpoints}{N}

\newcommand{\hdimset}{\Xi}
\newcommand{\hdimpt}[1]{\xi_{#1}}
\newcommand{\hdimpti}{\hdimpt{i}}
\newcommand{\hdimptj}{\hdimpt{j}}

\newcommand{\ldimset}{X}
\newcommand{\ldimpt}[1]{x_{#1}}
\newcommand{\ldimpti}{\ldimpt{i}}
\newcommand{\ldimptj}{\ldimpt{j}}

\newcommand{\hdimrank}[2]{\rho_{#1#2}}
\newcommand{\ldimrank}[2]{r_{#1#2}}

\newcommand{\hdimdist}[2]{\delta_{#1#2}}
\newcommand{\ldimdist}[2]{d_{#1#2}}